\documentclass{article}

\usepackage{microtype}
\usepackage{graphicx}
\usepackage{subfigure}
\usepackage{booktabs} %

\usepackage{hyperref}

\usepackage[accepted]{icml2018}

\icmltitlerunning{Generating Counterfactual Explanations with Natural Language}

\begin{document}

\twocolumn[
\icmltitle{Generating Counterfactual Explanations with Natural Language}

\begin{icmlauthorlist}
\icmlauthor{Lisa Anne Hendricks}{to}
\icmlauthor{Ronghang Hu}{to}
\icmlauthor{Trevor Darrell}{to}
\icmlauthor{Zeynep Akata}{goo}
\end{icmlauthorlist}

\icmlaffiliation{to}{University of California, Berkeley EECS}
\icmlaffiliation{goo}{University of Amsterdam}

\icmlcorrespondingauthor{Lisa Anne Hendricks}{lisa\_anne@berkeley.edu}

\vskip 0.3in
]

\printAffiliationsAndNotice{}  %

\begin{abstract}
Natural language explanations of deep neural network decisions 
provide an intuitive way for a AI agent to articulate a reasoning process.
Current textual explanations learn to discuss class discriminative features in an image. 
However, it is also helpful to understand which attributes might change a classification decision if present in an image (e.g., ``This is not a Scarlet Tanager because it does \textbf{not} have black wings.'')
We call such textual explanations \textit{counterfactual explanations}, and propose an intuitive method to generate counterfactual explanations by inspecting which evidence in an input is missing, but might contribute to a different classification decision if present in the image.
To demonstrate our method we consider a fine-grained image classification task in which we take as input an image and a counterfactual class and output text which explains why the image does not belong to a counterfactual class.
We then analyze our generated counterfactual explanations both qualitatively and quantitatively using proposed automatic metrics.
\end{abstract}

\section{Introduction}

Natural language is an intuitive and efficient way to communicate with AI agents about complex data, such as images.
Recently, natural language has been used to generate textual explanations which discuss why a decision made by a deep network is reasonable~\cite{hendricks2016generating,barratt2017interpnet,park2018multimodal}.
In this work, we consider textual explanations which reason about information which is not in an image, but might impact the decision if it were available. 
In this work, we refer to such explanations as \textit{counterfactual explanations}.
Such explanations can help provide helpful information to a human if a human is trying to discern why some input belongs to one class and not another~\cite{wachter2017counterfactual, dhurandhar2018explanations}.

Frequently, datasets used for textual explanation tasks include information about what kinds of discriminative attributes are present in a particular image.
However, data which discusses which information is not in an image, but might change the decision of a network if it were present, is not easily available.
Additionally, such data would be much harder to collect.
Instead of collecting one explanation for why an input belongs to a certain class,
explanations for why the input does not belong to all other possible classes would need to be collected instead.
Consequently, in this work, we do not have access to ground truth counterfactual explanations.
Thus, to generate counterfactual explanations, we reason that evidence which is discriminative for one class (class A), but not present in the image of another class (class B) can be considered counterfactual evidence which explains why the image does not belong to class A.
One way to generate counterfactual explanations would be to alter an input and observe how the output changes.
However, though there is prior work that considers perturbing images~\cite{grosse2016adversarial} and observing the changes in the predicted class, such perturbations are generally small and go un-noticed by the human eye.
To understand how a semantic feature (e.g., wing color) relates to a class label, it would be preferable to selectively change the color of the wing for an input image.
Performing such perturbations on natural images is challenging, so instead, we consider reasoning about counterfactuals in a semantic space. We determine which evidence is discriminative for a counterfactual class given text explanations for that class, then check if the evidence is in the image.
Like prior textual explanation work, our explanations are post-hoc; our goal is to generate informative text about an image, but our generated text is not meant to reflect the internal reasoning process of the neural network.
We demonstrate that our method results in good counterfactual explanations both qualitatively and quantitatively by considering proposed automatic metrics.

\begin{figure}[t]
\includegraphics[width=8cm]{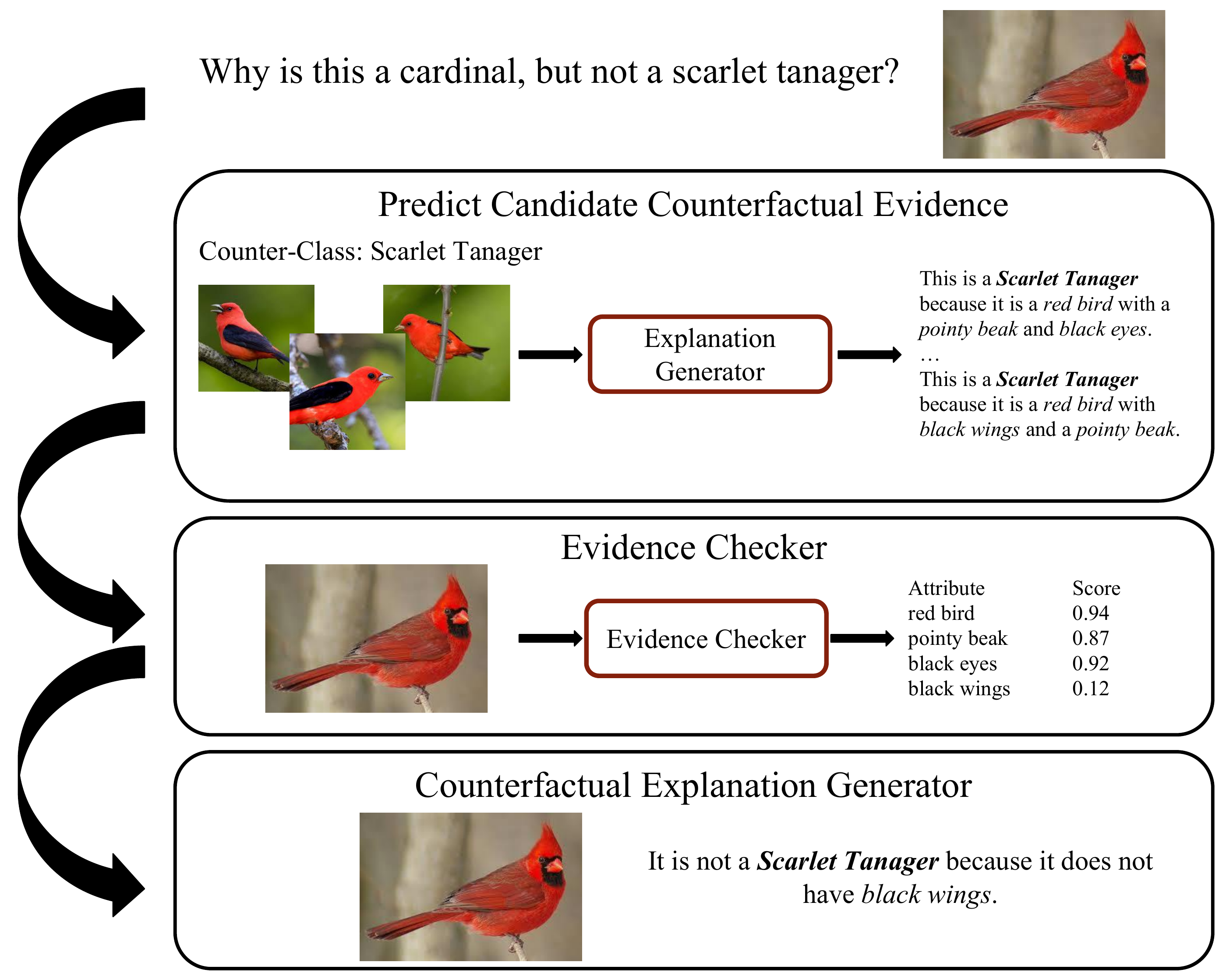}
\caption{Outline of our counterfactual explanation pipeline.  We first predict candidate counterfactual evidence, then determine if counterfactual evidence is in the image, then finally generates a cohesive sentence which mentions the counterfactual evidence which cannot be found in the image.}
\label{fig:method}
\end{figure}

\section{Generating Counter-Factual Explanations}

As an input to our system, we expect an image with a certain class label as well as another class label (we call this the counterfactual class or counter-class).
Our goal is to generate  counterfactual explanations which detail why image does not belong to the counter-class.

Figure~\ref{fig:method} outlines our approach for generating textual counterfactual explanations.
First, we use an explanation model to predict \textit{candidate counterfactual evidence}, or evidence which is discriminative for a counter-class.
We then verify if counterfactual evidence is in a given image using an \textit{evidence checker}.
Since we only have access to sentences which describe what is in an image, to generate counterfactual explanations, we negate phrases which do not appear in the image to generate a cohesive counterfactual explanation.
Below we detail how we perform each of these steps.

\subsection{Determining Candidate Counterfactual Evidence}
We first predict candidate counterfactual evidence given a counter-class.  To determine candidate counterfactaul evidence, we rely on the explanation model proposed in~\cite{hendricks2016generating}.
For a specific counter-class, we determine discriminative phrases by extracting noun phrases from generated explanations for images in the counter-class using a rule-based noun phrase chunker.
We consider all noun phrases present in explanations for the counter-class as candidate counterfactual evidence.

\subsection{Evidence Checker}  
Once candidate counterfactual evidence has been extracted, we verify which counterfactual evidence is \textit{not} present in the image using an evidence checker.
Candidate counterfactual evidence which is not present in the image can be considered counterfactual evidence and is used to generate our counterfactual explanations.
Unfortunately, we do not have access to groundtruth counterfactual evidence.
Instead we must rely on human descriptions of images, which indicate which attributes are in an image.
To learn how to determine whether counterfactual evidence is in an image, following~\cite{hendricks2017grounding}, we instead mine negative attributes by relying on the intuition that most visual attributes are exclusive, e.g., if an eye is a \textit{red eye}, it cannot also be a \textit{black eye}.
By considering such ``flipped'' attributes, we can build a set of attributes which appear in an image and a set of attributes which do not appear in an image.

We explore two evidence checker models: the first resembles a classifier model and takes an image and phrase as input and outputs a binary label indicating whether a phrase corresponds to the image (Counterfactual: Classifier or CF: Classifier) and the second is the phrase-critic architecture proposed in~\cite{hendricks2017grounding} (Counterfactual: Phrase-Critic or CF: Phrase-Critic).

\paragraph{Counterfactual: Classifier.}  The Counterfactual: Classifier first extracts visual features and textual features, then combines visual and textual features using elementwise multiplication.
We extract $conv_5$ from the model trained for fine-grained bird classification detailed in~\cite{hendricks2016generating} and text features using an LSTM.
Following \cite{park2018multimodal}, we combine vision and text modalities using elementwise multiplication and $L2$ normalization.
Once combining language and visual features, we apply a fully connected layer which predicts a score indicating whether or not a phrase is in the image (a score close to $0$ indicates the phrase is not in an image and a score close $1$ indicates the phrase is in an image).
Given a set of candidate counterfactual evidence, we determine which counterfactual evidence to discuss in our counterfacatul explanation by selecting the evidence with the \textit{minimum} score.

\paragraph{Counterfactual: Phrase-Critic.} As an alternative to simply classifying if a noun phrase applies to an image, we consider grounding, or localizing, natural language phrases in an image.
However, collecting bounding box annotations for natural language phrases can be difficult and the dataset we consider in this work does not have ground truth bounding box annotations.
Thus, following~\cite{hendricks2017grounding}, we employ an out-of-the-box grounding model, specifically the baseline grounding model from~\cite{hu2017modeling} which is trained on the Visual Genome dataset~\cite{krishna2017visual}.
As documented by~\cite{hendricks2017grounding}, employing out-of-the-box grounding models can be challenging because scores are not normalized to our dataset.
We thus train a phrase-critic model which takes as input an explanation and predicts a score indicating how well the explanation is grounded in the image, using the original sentence as a positive example and an example with ``flipped'' attributes as a negative example.
Unlike~\cite{hendricks2017grounding}, we use the phrase-critic to determine if phrases are \textit{not} present in an image.
Thus, to determine which counterfactual evidence is not in an image we input candidate counterfactual phrases into the phrase-critic model and select the counterfactual evidence which results in the \textit{minimum} phrase-critic score.

\subsection{Generating Counterfactual Text}

After determining candidate counterfactual evidence and determine possible counterfactual evidence with the evidence checker, we generate fluent textual explanations.
We use a rule-based negation system to negate noun phrases (e.g., counterfactual evidence like ``this bird has brown wings'' is negated to ``this bird does not have brown wings''). %
We then append the noun phrases to explanatory text which indicates which counter-class is being compared to (e.g., ``This is not a Scarlet Tanager because...'').

\section{Experiments}

\begin{figure*}[t]
\includegraphics[width=\linewidth]{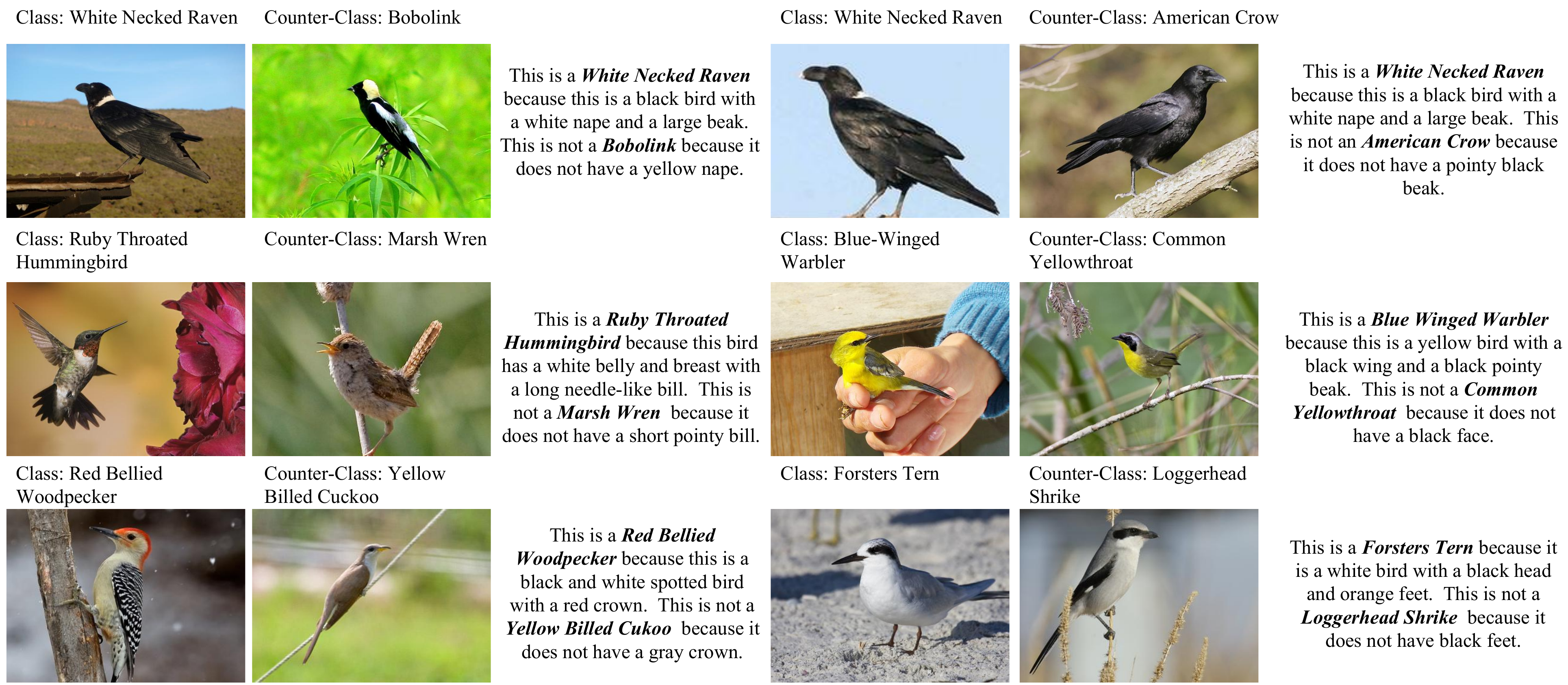}
\caption{Qualitative examples of counterfactual explanations.  Image on the left, with example image from counter-class on the right.  We show the explanation from~\cite{hendricks2016generating} followed by our generated counterfactual explanation.}
\label{fig:qual}
\end{figure*}

\paragraph{Dataset.}  We employ the Caltech-UCSD Birds 200-2011 dataset~\cite{wah2011caltech} and corresponding visual descriptions~\cite{reed2016learning}. %

\paragraph{Determining Counter-Classes.}  For our experiments, we determine a counter-classes by considering which image is closest to an input image in $fc_7$ space, but belongs to a different class.
By considering images which are similar in the $fc_7$ space, we ensure that our counter-classes are non-trivial to explain; generally at least one attribute will be shared between the image and corresponding counter-class.

\paragraph{Baseline.}  As a baseline, we randomly choose an attribute from our candidate counterfactual evidence as opposed to using a trained evidence checker.

\paragraph{Metrics.}  To measure the performance of counterfactual explanation models, we propose two metrics: phrase error and accuracy with counterfactual text.
Counterfactual text should not appear in the ground truth descriptions of an image.
We report if counterfactual text appears in ground truth descriptions of an image and refer to this as \textit{phrase error}.
Ideally, the phrase error should be 0.

Additionally, we observe how adding counterfactual information to a textual explanation impacts sentence-based classification.
\cite{hendricks2016generating} encourage generated text to be class discriminative by providing a positive reward to the sentence generation model at train time when generated sentences are discriminative.
To determine if generated sentences are discriminative, \cite{hendricks2016generating} employs a text based classifier which takes a sentence as input and outputs a class label.
Given an explanation for a class like cardinal (e.g., ``this is a red bird with a red beak''), the sentence classifier should predict the class ``cardinal''.
However, if counterfactual text is added to the explanation (e.g., ``this is a red bird with a red beak and green feathers''), the accuracy of the sentence classifier should decrease.
We call this metric \textit{accuracy with counterfactual text}.

\begin{table}
\centering
\begin{tabular}{ l|r|r } 
 \toprule
  & Phrase Error & Accuracy w/CF Text\\ 
  \hline
 Baseline & 16.26 & 39.54 \\ 
 CF: Classifier & 8.99 & 38.16  \\
 CF: Phrase-Critic & \textbf{7.37} & \textbf{36.62} \\
 \hline
\end{tabular}
\caption{Performance of different models using proposed metrics.  We consider phrase error and Accuracy w/Counterfactual (CF) Text.  Lower is better for both metrics.}
\label{tab:counterfactual}
\end{table}

\paragraph{Results.}  Table~\ref{tab:counterfactual} shows our quantitative evaluation on counterfactual explanations.
Considering the phrase error metric, we note that our baseline is quite strong, with an error of only 16.26\%.
However, both our CF: Classifier and CF: Phrase-Critic models outperform the baseline by a large margin.
Considering the Accuracy w/FC Test metric, all three models decrease the accuracy (accuracy without the counterfactual text is 54.38\%). 
However, we again note that our models outperform the baseline.

Interestingly, our classifier trained on the CUB data specifically for the task of determining if a phrase is in an image, performs worse than the phrase-critic model.
We hypothesize two reasons for this: (1) training on external data (in this case Visual Genome) is helpful for localizing fine-grained attributes and (2) localizing noun phrases, as opposed to just classifying whether or not they are in an image, is important.

Figure~\ref{fig:qual} shows example qualitative results. 
Figure~\ref{fig:qual} top left show an example which explains why the bird on the left (a ``White Necked Raven'') is not a ``Bobolink''.
The explanation from~\cite{hendricks2016generating} mentions the ``white nape'' of the bird, which is a discriminative feature for ``White Necked Raven''.  
In contrast, ``yellow nape'' is a discriminative feature for a similar bird class, ``Bobolink''. 
Our counterfactual explanation states that ``This is not a Bobolink because it does not have a yellow nape.''
Another discriminative attribute of the ``White Necked Raven'' is its distinctive blunt, thick bill.
When compared to the ``American Crow'' in Figure~\ref{fig:qual} (top right) our counterfactual explanation mentions that the white necked raven is not an ``American Crow'' because American Crows have a ``pointy black beak''.

Though the explanation model from~\cite{hendricks2016generating} should discuss discriminative bird features, sometimes the explanations are not discriminative for closely related birds.  
For example, in Figure~\ref{fig:qual} (middle right), the explanation from~\cite{hendricks2016generating} mentions that the bird is a ``Blue Winged Warbler'' because it is ``a yellow bird with a black wing and a black pointy beak.''  
This explanation could be applied to the example bird from the counter-class ``Common Yellowthroat''. 
However, when including the counterfactual explanation generated by the method described in this work (``This is not a Common Yellowthroat because it does not have a black face.''), it is immediately clear that the bird on the left must be a ``Blue-Winged Warbler'' and the bird on the right is a ``Common Yellowthroat.''
Yellow faces on yellow birds are common, so it is likely that models which just discuss discriminative features in an image do not learn to mention yellow faces on yellow birds.
In this case, the absence of a feature (``black face'') is helpful when explaining the bird category.

\bibliography{example_paper}
\bibliographystyle{icml2018}

\end{document}